\documentclass[10pt,twocolumn,letterpaper]{article}

\usepackage{cvpr}              

\usepackage{graphicx}
\usepackage{amsmath}
\usepackage{amssymb}
\usepackage{booktabs}

\usepackage[pagebackref,breaklinks,colorlinks]{hyperref}

\usepackage[capitalize]{cleveref}
\crefname{section}{Sec.}{Secs.}
\Crefname{section}{Section}{Sections}
\Crefname{table}{Table}{Tables}
\crefname{table}{Tab.}{Tabs.}

\def\cvprPaperID{*****} 
\def\confName{CVPR}
\def\confYear{2022}

\begin{document}

\title{Distilling Token-Pruned Pose Transformer for 2D Human Pose Estimation}

\author{Feixiang Ren\\
Sichuan University\\
{\tt\small angelfly.ren@hotmail.com}
}
\maketitle

\begin{abstract}



Human pose estimation has seen widespread use of transformer models in recent years. Pose transformers benefit from the self-attention map, which captures the correlation between human joint tokens and the image. However, training such models is computationally expensive. The recent token-Pruned Pose Transformer (PPT) solves this problem by pruning the background tokens of the image, which are usually less informative. However, although it improves efficiency, PPT inevitably leads to worse performance than TokenPose due to the pruning of tokens.

To overcome this problem, we present a novel method called  Distilling Pruned-Token Transformer for human pose estimation (DPPT). Our method leverages the output of a pre-trained TokenPose to supervise the learning process of PPT. We also establish connections between the internal structure of pose transformers and PPT, such as attention maps and joint features. Our experimental results on the MPII datasets show that our DPPT can significantly improve PCK compared to previous PPT models while still reducing computational complexity.

\end{abstract}

\section{Introduction}

Human pose estimation is a fundamental challenge in the field of computer vision that has been difficult to solve. The objective is to accurately locate human anatomical keypoints or body parts, such as the elbow, wrist, and so on. It has many applications, such as AR/VR, action recognition \cite{huang2017deep,yan2018spatial}, and medical diagnosis \cite{chen2021pd}. While deep convolutional neural networks (CNNs) have dominated the field of human pose estimation over the past few decades \cite{wei2016convolutional,newell2016stacked,xiao2018simple,sun2019deep,bestofboth,wang2020predicting}, they still struggle with occlusions when working with monocular images. 

In recent times, it has been well established that the transformer model \cite{vaswani2017attention,dosovitskiy2020image} can be highly effective in pose estimation tasks \cite{yuan2021tokens,yang2020transpose, ma2021transfusion,ma2022ppt, lin2020end}. While the convolution kernel in CNN only considers local correlation within adjacent pixels, the self-attention module of the transformer can establish global dependencies among all pixels in an image.
In the domain of single-view 2D human pose estimation, the transformer-based models TransPose \cite{yang2020transpose} and TokenPose \cite{li2021tokenpose} have gained prominence. The model can effectively learn the dependencies within the features as well as across the joint tokens and feature map through the attention map, thereby improving accuracy. 
However, the computation complexity of dense attention is quadratic to the number of tokens, which makes it difficult to scale up to high-resolution feature maps. 
To address this issue, the PPT \cite{ma2022ppt} proposed a method to accelerate TokenPose by removing the feature tokens that have relatively lower attention values, with a motivation that background tokens cannot provide effective information or even introduce noise. 
Nonetheless, it is worth noting that PPT gains efficiency at the cost of accuracy, as the pruned tokens may still contain useful information in both training and inference time.

In the present study, we introduce a new approach called Distilling-PPT (DPPT) to enhance the efficiency of TokenPose while maintaining its accuracy. DPPT utilizes a pre-trained TokenPose to facilitate PPT's learning process. To achieve this, we adopt knowledge distillation (KD) \cite{hinton2015distilling} to formulate PPT's learning, i.e., treating the pre-trained TokenPose as the teacher model and PPT as the student model. It is worth noting that TokenPose and PPT possess identical parameter counts, with the only difference being the number of processed tokens in their transformer encoder layers. 
Thus, this is a type of self-distillation, rather than a traditional setting of KD that transfers the ability of a complicated network to a lightweight network. 
Besides the output heatmaps of TokenPose, several intermediate features of TokenPose, such as joint tokens and attention maps, can serve as natural guidance. Our experimental results indicate that DPPT yields superior accuracy performance compared to PPT at the same sparsity level. Therefore, our approach provides a more favorable balance between efficiency and accuracy than PPT.


Our main contributions are summarized as follows:
\begin{enumerate}
    
    \item We propose the Distilling-PPT for efficient 2D human pose estimation by training a PPT with the guidance of pre-trained TokenPose.
    
    \item We evaluate several types of distilling and justify what kinds of information from the teacher are helpful. 
    
    \item The experimental results on MPII indicate that our DPPT can achieve higher accuracy than PPT at the same level of efficiency.
    
\end{enumerate}

\section{Related Work}

\subsection{Pose Estimation with Transformers}
In the field of pose estimation, significant improvements have been made to convolutional neural networks (CNNs) over the past few years through the introduction of multi-scale representations \cite{chen2018cascaded,newell2016stacked,chu2017multi,xiao2018simple,sun2019deep} and the use of skeletal structure \cite{tompson2014joint,kong2019adaptive,kong2020rotation,chen2020nonparametric,kong2020sia}.
Despite its effectiveness, the convolutional approach still faces challenges in capturing and modeling long-range relationships due to its inherent locality nature. This is particularly challenging in pose estimation tasks, where long-range dependencies are critical for accurate results. 

Recently, transformers have gained attention in the field of pose estimation \cite{yang2020transpose,li2021tokenpose,mao2021tfpose,li2021pose,lin2020end,ma2022ppt, xu2022vitpose} due to their ability to capture global features.
For instance, TransPose \cite{yang2020transpose} applies transformers on CNN feature maps to explain keypoint correspondence. 
TokenPose \cite{li2021tokenpose} utilizes ViT\cite{dosovitskiy2020image} as a foundation and includes additional keypoint tokens to learn the relationships between different joints and visual cues. 
The efficiency of TokenPose has been further enhanced by the recent introduction of PPT \cite{ma2022ppt}, which prunes unnecessary background tokens based on the attention maps of keypoint tokens.
However, PPT directly trains the model from scratch, which may easily identify background tokens by mistake.

\subsection{Efficient Vision Transformers}

The transformer \cite{vaswani2017attention} has recently made significant advancements in various computer vision tasks, including classification\cite{dosovitskiy2020image,touvron2020training}, object detection\cite{carion2020end,zhu2020deformable,fang2021you, dai2021up}, semantic segmentation \cite{zheng2020rethinking,strudel2021segmenter,yan2022after}, and pose estimation \cite{yang2020transpose,li2021tokenpose}. 
Despite its promising accuracy, the ViT \cite{dosovitskiy2020image} is a computational heavyweight. To address this issue, several algorithms have been proposed to improve the efficiency of vision transformers in different ways \cite{liu2021swin,chen2021chasing,yu2022unified,chen2022dearkd,shen2020q,sun2022vaqf,kong2022peeling}.

Among them, token pruning \cite{yuan2021tokens,caron2021emerging,ryoo2021tokenlearner,rao2021dynamicvit,kong2021spvit,liang2022evit} is an effective way to speed up the transformer as it does not alter the network architecture. To accomplish this, less important tokens can be determined through training additional token selectors \cite{ryoo2021tokenlearner, rao2021dynamicvit} or directly using attention weights \cite{liang2022evit,kong2021spvit}. These methods have shown considerable potential in improving the computational efficiency of vision transformers while maintaining competitive performance. 
PPT \cite{ma2022ppt} also follows the design of EviT \cite{liang2022evit} that uses the attention matrix of special tokens to find important tokens.

\subsection{Knowledge Distillation}
Knowledge distillation (KD) \cite{hinton2015distilling} offers a straightforward approach to enhance the performance of a compact model (students) by acquiring knowledge from a pre-trained model (teachers) that has much more complex networks.
The knowledge of the teacher can be obtained from the output probabilities (logits) \cite{mirzadeh2019improved,zhang2019your} and intermediate feature maps \cite{passalis2018learning, ahn2019variational}. 
Besides learning from a complicated network, students can also learn from the pre-trained version of itself (a.k.a, self-KD) \cite{yuan2020revisiting}. 
Recently, several works suggest that KD can be applied to vision transformers \cite{chen2022dearkd,lin2022knowledge,touvron2020training}. 
In the pose estimation area, previous works also show that KD plays an important role in efficient human pose estimation with CNNs backbones \cite{zhang2019fast,li2021online}.

\section{Methodology}

\subsection{Revisiting Token Pruning in Pose Transformer}
The study by Ma et al. \cite{ma2022ppt} indicates that a gradual reduction in the number of visual tokens used in TokenPose \cite{yuan2021tokens} can enhance its efficiency while maintaining accuracy.
The procedure involves partitioning the 2D feature map $F$ of an image into patches, which are then transformed into a 1D sequence of visual tokens $X_p \in \mathbb{R}^{N_v \times D}$ using a linear projector. The resulting visual tokens are then concatenated with $J$ learnable keypoint tokens $X_k \in \mathbb{R}^{J \times D}$ to form the input sequence $X^0 = [X_k, X_v] \in \mathbb{R}^{N \times D}$ for the transformer, where $N = N_v + J$.

PPT incorporates the human token identification (HTI) module to extract $r \times N_v$ foreground tokens based on the attention map $A\in \mathbb{R}^{J\times N_v}$ of the keypoint tokens. The remaining tokens are pruned. In this context, $r$ denotes the retention ratio. Subsequently, the length of the sequence is reduced to $rN_v + J$ for the subsequent transformer layers. HTI is inserted into the transformer encoder layers $e$ times, resulting in the retention of only $r^e N_v + J$ tokens. Lastly, an MLP layer is applied to the keypoint tokens to obtain the output heatmaps $H \in \mathbf{R}^{J\times H \times W}$.  
However, how to locate the informative token at the early training stage is still challenging for PPT. Thus, PPT inevitably removes some important tokens unintentionally.

\begin{figure*}
    \centering
    \includegraphics[width=1.0\linewidth]{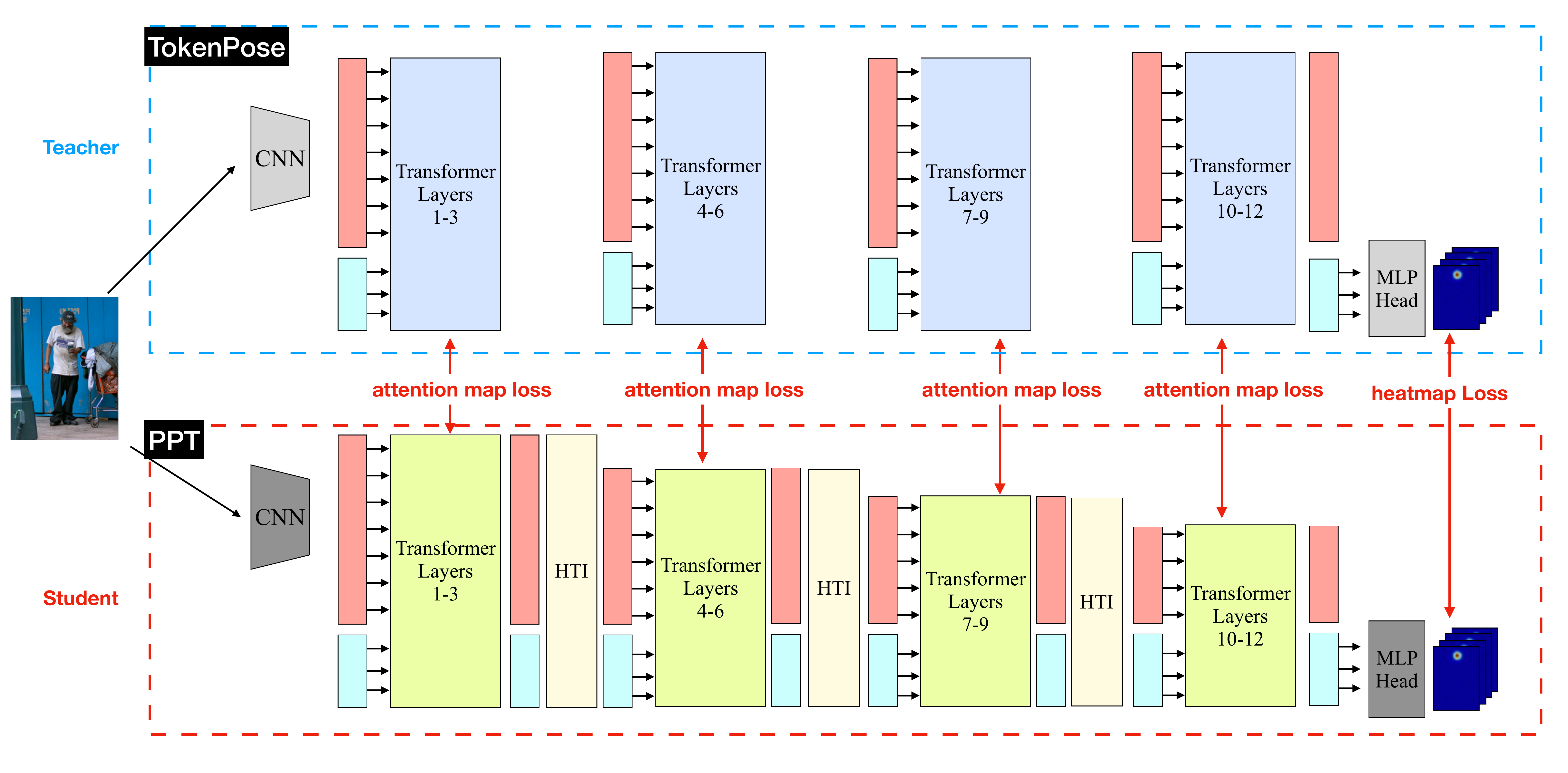}
    \caption{Overall architecture of Distilling-PPT. We use a pre-trained TokenPose as the teacher model and the pruned-token transformer (PPT) as the student networks. The student learns from both output heatmaps and attention maps from the teacher. }
    \label{fig:framework}
\end{figure*}

\subsection{Distilling Pruned-Token Transformer}

\paragraph{Overview} 
In our Distilling-PPT (DPPT), we  utilize a well-pre-trained TokenPose model to guide the learning process of PPT. 
For PPT, we use the same architecture as TokenPose. Thus, it can be considered as a self-distillation as both teacher and student has the same number of parameters. 
As shown in Figure \ref{fig:framework}, we consider the pre-trained TokenPose $f_T(\cdot)$ as the teacher model and the PPT model $f_P(\cdot)$ as the student model. 
We consider two types of guidance from the teacher model. One is the output heatmap, the other is the intermediate attention map. The former can provide a soft target than the ground truth heatmap, which contains dark knowledge learned by the teacher model \cite{hinton2015distilling}. The latter can help the student model locate important tokens correctly at the early training stage.

\paragraph{Implementation}
In detail, given an RGB image $I$, the standard training of PPT uses the meaning square loss $L_{gt} = (f_S(I) - Y)^2$ between the output from the students and the ground truth heatmap $Y$. In DPPT, we obtain the heatmap $H_T = f_T(I)$ from the pre-trained TokenPose as well. Meanwhile, we maintain the intermediate attention maps $A_T^l \in \mathbb{R}^{J\times N_v}$ of keypoint tokens from teacher models, where $l$ is the index of transformer encoder layers. 
We make the output of the student $f_S(I)$ to be similar to $H_T$ with $L_{hm}$:  
\begin{equation}
    L_{hm} = (f_S(I) - f_T(I))^2 
\end{equation}
Moreover, we enable the intermediate attention map of student $A_S^l$ to be similar to $A_T^l$ with a loss $L_{attn}$, which enables the student to select informative tokens. 
\begin{equation}
    L_{attn} = \sum_l (A_S^l - A_T^l)^2
\end{equation}
In summary, besides learning from the label $Y$, the student network is trained by
\begin{equation}
    L = L_{gt} + \lambda_1 L_{hm} + \lambda_2 L_{attn}
\end{equation}
In our experiment, we set both $\lambda_1$ and $\lambda_2$ to $1$ by default.

\section{Experiments}
\subsection{Experimental Setup}
\paragraph{Dataset} 
Our assessment of DPPT is conducted on the MPII dataset \cite{andriluka14cvpr}, comprising approximately $25K$ images and $40K$ instances of human beings, each annotated with $16$ keypoints. The evaluation metric used is the probability of correct keypoint (PCKh) score \cite{andriluka14cvpr}, which is normalized based on the head position. A keypoint is deemed accurate if its location is within a pre-defined threshold of the ground truth. We follow the convention and use PCKh@0.5.

\paragraph{Trainings}
We adopted the training process outlined in \cite{li2021tokenpose}. Specifically, all networks were optimized using the Adam optimizer \cite{kingma2014adam} and trained for $300$ epochs with Mean Square Error (MSE) loss. The learning rate was initialized to $0.001$ and reduced by a factor of $0.1$ at the $200$-th and $260$-th epochs. 
Instead of gradually reducing the keep ratio from $1$ to $r$ in \cite{ma2022ppt}, we directly train the networks with a fixed pruned keep ratio during the entire training.

\begin{table*}[!t]
\centering
\resizebox{1.0\textwidth}{!}{
\begin{tabular}{l|l|l|lllllll|c}
\toprule
Method & \#Params & GFLOPs & Head & Sho & Elb & Wri & Hip & Kne & Ank & Mean  \\
\hline
SimpleBaseline-Res50 \cite{xiao2018simple}  & 34M & 12.0 & 96.4 & 95.3 & 89.0 & 83.2 & 88.4 & 84.0 & 79.6 & 88.5  \\
SimpleBaseline-Res101 \cite{xiao2018simple} & 53M & 16.5 & 96.9 & 95.9 & 89.5 & 84.4 & 88.4 & 84.5 & 80.7 & 89.1 \\
SimpleBaseline-Res152 \cite{xiao2018simple} & 53M & 21.0 & 97.0 & 95.9 & 90.0 & 85.0 & 89.2 & 85.3 & 81.3 & 89.6 \\
HRNet-W32. \cite{sun2019deep}               & 28.5M & 9.5 & 96.9 & 96.0 & 90.6 & 85.8 & 88.7 & 86.6 & 82.6 & 90.1 \\
\hline
TokenPose-S \cite{li2021tokenpose}   & 7.7M  & 2.5 & 96.0 & 94.5 & 86.5 & 79.7 & 86.7 & 80.1 & 75.2 & 86.2 \\
PPT-S                                & 7.7M  & 1.9  & 96.6 & 94.9 & 87.6 & 81.3 & 87.1 & 82.4 & 76.7 & 87.3 \\
DPPT-S &  7.7M  & 1.9  & 96.4 & 94.9 & 88.3 & 81.8 &  88.2 &  83.0 &  78.3 &  \textbf{87.9} \\
\bottomrule
\end{tabular}
}
\caption{{Results on the MPII (PCKh@0.5) withinput size is $256\times256$.}}
\label{tab:mpii_2d}
\end{table*}

\paragraph{Implementation Details}
To ensure a fair comparison with PPT \cite{ma2022ppt}, we trained PPT-S with the guidance of TokenPose-S \cite{li2021tokenpose}. We denote it as DPPT-S. Thus, it is a kind of self-distillation. 
Specifically, in the case of PPT-S, we set the number of encoder layers $L$ to $12$, embedding size $D$ to $192$, and the number of heads $H$ to $8$, while utilizing the shallow stem-net as the CNN backbone. Following the approach taken in PPT\cite{ma2022ppt}, we inserted the token pruning module before the $4^{th}$, $7^{th}$, and $10^{th}$ encoder layers.
For PPT-L/D6, we adopted HRNet-W48 as the backbone and inserted the token pruning module before the $2^{th}$, $4^{th}$, and $5^{th}$ encoder layers. 
Similar to the previous network, we used $12$ encoder layers and set $256$ as the number of visual tokens $N_v$ for all networks.

\subsection{Results}
The outcomes are presented in Table \ref{tab:mpii_2d}. It is evident from the results that DPPT surpasses PPT\cite{ma2022ppt} on various architectures. Notably, DPPT-S exhibits a $1.7\%$ improvement over TokenPose, which is $0.6\%$ higher than PPT-S. Moreover, the performance enhancement on challenging keypoints, such as the elbow and ankle, is even more significant (i.e., from 87.6 to 88.3 on the elbow and from 76.7 to 78.3 on the ankle). 
Note that, these groups of keypoints usually have more freedom of movement. 
Additionally, DPPT-S reduces FLOPs by $24\%$, compared to PPT-S.
Therefore, leveraging pre-trained Tokenpose can effectively enhance the performance of PPT while preserving its efficiency.

\subsection{Ablation Studies}
To assess the significance of each distilling loss, we conducted additional ablation studies. Specifically, we evaluated the performance of DPPT when training with only $L_{hm}$ and only $L_{gt} + L_{hm}$, in addition to the default training approach.
The results presented in Table \ref{tab:ablation} indicate that removing $L_{attn}$ causes a drop in PCK from $87.9$ to $87.3$, highlighting the importance of the attention map guidance from the teacher model (TokenPose). This guidance aids the student model (PPT) in accurately identifying foreground tokens.
Furthermore, removing $L_{hm}$ results in a PCK drop from $87.3$ to $86.7$, emphasizing the importance of learning from the output heatmaps of the teacher model, which serves as a soft target. 
It is worth noting that PPT employs a warm-up strategy that gradually reduces the keep ratio from $1.0$ to $r$ during training, which necessitates ad-hoc hyperparameter selection. In contrast, our approach uses a fixed keep ratio throughout training without hyperparameter tuning. Thus, the performance achieved using $L_{gt}$ only (86.7) is slightly worse than PPT (87.3). 

\begin{table}[h]
    \centering
    \resizebox{0.75\linewidth}{!}{
    \begin{tabular}{l|c}
    \toprule
         Loss functions & PCKs@0.5 \\
         \hline 
         $L_{gt}$  & 86.7 \\
          $L_{gt} + L_{hm}$ & 87.3 \\
         $L_{gt} + L_{hm} + L_{attn}$   & 87.9 \\
    \bottomrule  
    \end{tabular}
    }
    \caption{Ablation studies on different loss. }
    \label{tab:ablation}
\end{table}

\section{Conclusion}

This paper presents a novel approach called Distilling-PPT, which incorporates self-distillation into the Pruned-Token Pose Transformer (PPT) \cite{ma2022ppt} to enhance the efficiency of TokenPose \cite{li2021tokenpose} for pose estimation. Specifically, we leverage the pre-trained TokenPose as the teacher model and the PPT as the lightweight student model to improve the performance of PPT. Our experimental results demonstrate that Distilling-PPT outperforms both PPT and TokenPose while maintaining the same speedup ratio as PPT. These findings are expected to aid the research community in developing more efficient transformer architectures for pose estimation.

{\small
\bibliographystyle{ieee_fullname}
\bibliography{egbib}

\begin{thebibliography}{10}\itemsep=-1pt

\bibitem{ahn2019variational}
Sungsoo Ahn, Shell~Xu Hu, Andreas Damianou, Neil~D Lawrence, and Zhenwen Dai.
\newblock Variational information distillation for knowledge transfer.
\newblock In {\em CVPR}, 2019.

\bibitem{andriluka14cvpr}
Mykhaylo Andriluka, Leonid Pishchulin, Peter Gehler, and Bernt Schiele.
\newblock 2d human pose estimation: New benchmark and state of the art
  analysis.
\newblock In {\em CVPR}, 2014.

\bibitem{carion2020end}
Nicolas Carion, Francisco Massa, Gabriel Synnaeve, Nicolas Usunier, Alexander
  Kirillov, and Sergey Zagoruyko.
\newblock End-to-end object detection with transformers.
\newblock In {\em ECCV}, 2020.

\bibitem{caron2021emerging}
Mathilde Caron, Hugo Touvron, Ishan Misra, Herv{\'e} J{\'e}gou, Julien Mairal,
  Piotr Bojanowski, and Armand Joulin.
\newblock Emerging properties in self-supervised vision transformers.
\newblock In {\em ICCV}, 2021.

\bibitem{chen2021chasing}
Tianlong Chen, Yu Cheng, Zhe Gan, Lu Yuan, Lei Zhang, and Zhangyang Wang.
\newblock Chasing sparsity in vision transformers: An end-to-end exploration.
\newblock {\em NeurIPS}, 2021.

\bibitem{chen2022dearkd}
Xianing Chen, Qiong Cao, Yujie Zhong, Jing Zhang, Shenghua Gao, and Dacheng
  Tao.
\newblock Dearkd: Data-efficient early knowledge distillation for vision
  transformers.
\newblock In {\em CVPR}, 2022.

\bibitem{chen2020nonparametric}
Yifei Chen, Haoyu Ma, Deying Kong, Xiangyi Yan, Jianbao Wu, Wei Fan, and
  Xiaohui Xie.
\newblock Nonparametric structure regularization machine for 2d hand pose
  estimation.
\newblock In {\em WACV}, 2020.

\bibitem{chen2021pd}
Yifei Chen, Haoyu Ma, Jiangyuan Wang, Jianbao Wu, Xian Wu, and Xiaohui Xie.
\newblock Pd-net: Quantitative motor function evaluation for parkinson's
  disease via automated hand gesture analysis.
\newblock In {\em KDD}, 2021.

\bibitem{chen2018cascaded}
Yilun Chen, Zhicheng Wang, Yuxiang Peng, Zhiqiang Zhang, Gang Yu, and Jian Sun.
\newblock Cascaded pyramid network for multi-person pose estimation.
\newblock In {\em CVPR}, 2018.

\bibitem{chu2017multi}
Xiao Chu, Wei Yang, Wanli Ouyang, Cheng Ma, Alan~L Yuille, and Xiaogang Wang.
\newblock Multi-context attention for human pose estimation.
\newblock In {\em CVPR}, 2017.

\bibitem{dai2021up}
Zhigang Dai, Bolun Cai, Yugeng Lin, and Junying Chen.
\newblock Up-detr: Unsupervised pre-training for object detection with
  transformers.
\newblock In {\em Proceedings of the IEEE/CVF conference on computer vision and
  pattern recognition}, pages 1601--1610, 2021.

\bibitem{dosovitskiy2020image}
Alexey Dosovitskiy, Lucas Beyer, Alexander Kolesnikov, Dirk Weissenborn,
  Xiaohua Zhai, Thomas Unterthiner, Mostafa Dehghani, Matthias Minderer, Georg
  Heigold, Sylvain Gelly, et~al.
\newblock An image is worth 16x16 words: Transformers for image recognition at
  scale.
\newblock {\em ICLR}, 2021.

\bibitem{fang2021you}
Yuxin Fang, Bencheng Liao, Xinggang Wang, Jiemin Fang, Jiyang Qi, Rui Wu,
  Jianwei Niu, and Wenyu Liu.
\newblock You only look at one sequence: Rethinking transformer in vision
  through object detection.
\newblock {\em NeurIPS}, 2021.

\bibitem{hinton2015distilling}
Geoffrey Hinton, Oriol Vinyals, and Jeff Dean.
\newblock Distilling the knowledge in a neural network.
\newblock {\em arXiv preprint arXiv:1503.02531}, 2015.

\bibitem{huang2017deep}
Zhiwu Huang, Chengde Wan, Thomas Probst, and Luc Van~Gool.
\newblock Deep learning on lie groups for skeleton-based action recognition.
\newblock In {\em CVPR}, 2017.

\bibitem{kingma2014adam}
Diederik~P Kingma and Jimmy Ba.
\newblock Adam: A method for stochastic optimization.
\newblock {\em ICLR}, 2015.

\bibitem{kong2019adaptive}
Deying Kong, Yifei Chen, Haoyu Ma, Xiangyi Yan, and Xiaohui Xie.
\newblock Adaptive graphical model network for 2d handpose estimation.
\newblock {\em BMVC}, 2019.

\bibitem{kong2020rotation}
Deying Kong, Haoyu Ma, Yifei Chen, and Xiaohui Xie.
\newblock Rotation-invariant mixed graphical model network for 2d hand pose
  estimation.
\newblock In {\em WACV}, 2020.

\bibitem{kong2020sia}
Deying Kong, Haoyu Ma, and Xiaohui Xie.
\newblock Sia-gcn: A spatial information aware graph neural network with 2d
  convolutions for hand pose estimation.
\newblock {\em BMVC}, 2020.

\bibitem{kong2021spvit}
Zhenglun Kong, Peiyan Dong, Xiaolong Ma, Xin Meng, Wei Niu, Mengshu Sun, Bin
  Ren, Minghai Qin, Hao Tang, and Yanzhi Wang.
\newblock Spvit: Enabling faster vision transformers via soft token pruning.
\newblock {\em ECCV}, 2022.

\bibitem{kong2022peeling}
Zhenglun Kong, Haoyu Ma, Geng Yuan, Mengshu Sun, Yanyue Xie, Peiyan Dong, Xin
  Meng, Xuan Shen, Hao Tang, Minghai Qin, et~al.
\newblock Peeling the onion: Hierarchical reduction of data redundancy for
  efficient vision transformer training.
\newblock {\em AAAI}, 2023.

\bibitem{li2021pose}
Ke Li, Shijie Wang, Xiang Zhang, Yifan Xu, Weijian Xu, and Zhuowen Tu.
\newblock Pose recognition with cascade transformers.
\newblock {\em CVPR}, 2021.

\bibitem{li2021tokenpose}
Yanjie Li, Shoukui Zhang, Zhicheng Wang, Sen Yang, Wankou Yang, Shu-Tao Xia,
  and Erjin Zhou.
\newblock Tokenpose: Learning keypoint tokens for human pose estimation.
\newblock In {\em ICCV}, 2021.

\bibitem{li2021online}
Zheng Li, Jingwen Ye, Mingli Song, Ying Huang, and Zhigeng Pan.
\newblock Online knowledge distillation for efficient pose estimation.
\newblock In {\em CVPR}, 2021.

\bibitem{liang2022evit}
Youwei Liang, Chongjian GE, Zhan Tong, Yibing Song, Jue Wang, and Pengtao Xie.
\newblock {EV}it: Expediting vision transformers via token reorganizations.
\newblock In {\em ICLR}, 2022.

\bibitem{lin2020end}
Kevin Lin, Lijuan Wang, and Zicheng Liu.
\newblock End-to-end human pose and mesh reconstruction with transformers.
\newblock {\em CVPR}, 2021.

\bibitem{lin2022knowledge}
Sihao Lin, Hongwei Xie, Bing Wang, Kaicheng Yu, Xiaojun Chang, Xiaodan Liang,
  and Gang Wang.
\newblock Knowledge distillation via the target-aware transformer.
\newblock In {\em CVPR}, 2022.

\bibitem{liu2021swin}
Ze Liu, Yutong Lin, Yue Cao, Han Hu, Yixuan Wei, Zheng Zhang, Stephen Lin, and
  Baining Guo.
\newblock Swin transformer: Hierarchical vision transformer using shifted
  windows.
\newblock In {\em ICCV}, 2021.

\bibitem{ma2021transfusion}
Haoyu Ma, Liangjian Chen, Deying Kong, Zhe Wang, Xingwei Liu, Hao Tang, Xiangyi
  Yan, Yusheng Xie, Shih-Yao Lin, and Xiaohui Xie.
\newblock Transfusion: Cross-view fusion with transformer for 3d human pose
  estimation.
\newblock {\em BMVC}, 2021.

\bibitem{ma2022ppt}
Haoyu Ma, Zhe Wang, Yifei Chen, Deying Kong, Liangjian Chen, Xingwei Liu,
  Xiangyi Yan, Hao Tang, and Xiaohui Xie.
\newblock Ppt: token-pruned pose transformer for monocular and multi-view human
  pose estimation.
\newblock In {\em ECCV}, 2022.

\bibitem{mao2021tfpose}
Weian Mao, Yongtao Ge, Chunhua Shen, Zhi Tian, Xinlong Wang, and Zhibin Wang.
\newblock Tfpose: Direct human pose estimation with transformers.
\newblock {\em arXiv preprint arXiv:2103.15320}, 2021.

\bibitem{mirzadeh2019improved}
Seyed-Iman Mirzadeh, Mehrdad Farajtabar, Ang Li, Nir Levine, Akihiro Matsukawa,
  and Hassan Ghasemzadeh.
\newblock Improved knowledge distillation via teacher assistant.
\newblock {\em AAAI}, 2020.

\bibitem{newell2016stacked}
Alejandro Newell, Kaiyu Yang, and Jia Deng.
\newblock Stacked hourglass networks for human pose estimation.
\newblock In {\em ECCV}, 2016.

\bibitem{passalis2018learning}
Nikolaos Passalis and Anastasios Tefas.
\newblock Learning deep representations with probabilistic knowledge transfer.
\newblock In {\em ECCV}, pages 268--284, 2018.

\bibitem{rao2021dynamicvit}
Yongming Rao, Wenliang Zhao, Benlin Liu, Jiwen Lu, Jie Zhou, and Cho-Jui Hsieh.
\newblock Dynamicvit: Efficient vision transformers with dynamic token
  sparsification.
\newblock {\em NeurIPS}, 2021.

\bibitem{ryoo2021tokenlearner}
Michael Ryoo, AJ Piergiovanni, Anurag Arnab, Mostafa Dehghani, and Anelia
  Angelova.
\newblock Tokenlearner: Adaptive space-time tokenization for videos.
\newblock {\em NeurIPS}, 2021.

\bibitem{shen2020q}
Sheng Shen, Zhen Dong, Jiayu Ye, Linjian Ma, Zhewei Yao, Amir Gholami,
  Michael~W Mahoney, and Kurt Keutzer.
\newblock Q-bert: Hessian based ultra low precision quantization of bert.
\newblock In {\em AAAI}, 2020.

\bibitem{strudel2021segmenter}
Robin Strudel, Ricardo Garcia, Ivan Laptev, and Cordelia Schmid.
\newblock Segmenter: Transformer for semantic segmentation.
\newblock In {\em Proceedings of the IEEE/CVF international conference on
  computer vision}, pages 7262--7272, 2021.

\bibitem{sun2019deep}
Ke Sun, Bin Xiao, Dong Liu, and Jingdong Wang.
\newblock Deep high-resolution representation learning for human pose
  estimation.
\newblock In {\em CVPR}, 2019.

\bibitem{sun2022vaqf}
Mengshu Sun, Haoyu Ma, Guoliang Kang, Yifan Jiang, Tianlong Chen, Xiaolong Ma,
  Zhangyang Wang, and Yanzhi Wang.
\newblock Vaqf: Fully automatic software-hardware co-design framework for
  low-bit vision transformer.
\newblock {\em arXiv preprint arXiv:2201.06618}, 2022.

\bibitem{tompson2014joint}
Jonathan~J Tompson, Arjun Jain, Yann LeCun, and Christoph Bregler.
\newblock Joint training of a convolutional network and a graphical model for
  human pose estimation.
\newblock {\em NIPS}, 2014.

\bibitem{touvron2020training}
Hugo Touvron, Matthieu Cord, Matthijs Douze, Francisco Massa, Alexandre
  Sablayrolles, and Herv{\'e} J{\'e}gou.
\newblock Training data-efficient image transformers \& distillation through
  attention.
\newblock {\em ICML}, 2021.

\bibitem{vaswani2017attention}
Ashish Vaswani, Noam Shazeer, Niki Parmar, Jakob Uszkoreit, Llion Jones,
  Aidan~N Gomez, Lukasz Kaiser, and Illia Polosukhin.
\newblock Attention is all you need.
\newblock {\em NIPS}, 2017.

\bibitem{wang2020predicting}
Zhe Wang, Daeyun Shin, and Charless~C Fowlkes.
\newblock Predicting camera viewpoint improves cross-dataset generalization for
  3d human pose estimation.
\newblock In {\em ECCV 3DPW workshop}, 2020.

\bibitem{bestofboth}
Z. Wang, J. Yang, and C. Fowlkes.
\newblock The best of both worlds: Combining model-based and nonparametric
  approaches for 3d human body estimation.
\newblock In {\em CVPR ABAW workshop}, 2022.

\bibitem{wei2016convolutional}
Shih-En Wei, Varun Ramakrishna, Takeo Kanade, and Yaser Sheikh.
\newblock Convolutional pose machines.
\newblock In {\em CVPR}, 2016.

\bibitem{xiao2018simple}
Bin Xiao, Haiping Wu, and Yichen Wei.
\newblock Simple baselines for human pose estimation and tracking.
\newblock In {\em ECCV}, 2018.

\bibitem{xu2022vitpose}
Yufei Xu, Jing Zhang, Qiming Zhang, and Dacheng Tao.
\newblock Vi{TP}ose: Simple vision transformer baselines for human pose
  estimation.
\newblock In {\em NeurIPS}, 2022.

\bibitem{yan2018spatial}
Sijie Yan, Yuanjun Xiong, and Dahua Lin.
\newblock Spatial temporal graph convolutional networks for skeleton-based
  action recognition.
\newblock In {\em AAAI}, 2018.

\bibitem{yan2022after}
Xiangyi Yan, Hao Tang, Shanlin Sun, Haoyu Ma, Deying Kong, and Xiaohui Xie.
\newblock After-unet: Axial fusion transformer unet for medical image
  segmentation.
\newblock In {\em WACV}, 2022.

\bibitem{yang2020transpose}
Sen Yang, Zhibin Quan, Mu Nie, and Wankou Yang.
\newblock Transpose: Keypoint localization via transformer.
\newblock In {\em ICCV}, 2021.

\bibitem{yu2022unified}
Shixing Yu, Tianlong Chen, Jiayi Shen, Huan Yuan, Jianchao Tan, Sen Yang, Ji
  Liu, and Zhangyang Wang.
\newblock Unified visual transformer compression.
\newblock In {\em ICLR}, 2022.

\bibitem{yuan2021tokens}
Li Yuan, Yunpeng Chen, Tao Wang, Weihao Yu, Yujun Shi, Zi-Hang Jiang,
  Francis~EH Tay, Jiashi Feng, and Shuicheng Yan.
\newblock Tokens-to-token vit: Training vision transformers from scratch on
  imagenet.
\newblock In {\em ICCV}, 2021.

\bibitem{yuan2020revisiting}
Li Yuan, Francis~EH Tay, Guilin Li, Tao Wang, and Jiashi Feng.
\newblock Revisiting knowledge distillation via label smoothing regularization.
\newblock In {\em CVPR}, 2020.

\bibitem{zhang2019fast}
Feng Zhang, Xiatian Zhu, and Mao Ye.
\newblock Fast human pose estimation.
\newblock In {\em CVPR}, 2019.

\bibitem{zhang2019your}
Linfeng Zhang, Jiebo Song, Anni Gao, Jingwei Chen, Chenglong Bao, and Kaisheng
  Ma.
\newblock Be your own teacher: Improve the performance of convolutional neural
  networks via self distillation.
\newblock In {\em ICCV}, 2019.

\bibitem{zheng2020rethinking}
Sixiao Zheng, Jiachen Lu, Hengshuang Zhao, Xiatian Zhu, Zekun Luo, Yabiao Wang,
  Yanwei Fu, Jianfeng Feng, Tao Xiang, Philip~HS Torr, et~al.
\newblock Rethinking semantic segmentation from a sequence-to-sequence
  perspective with transformers.
\newblock {\em CVPR}, 2021.

\bibitem{zhu2020deformable}
Xizhou Zhu, Weijie Su, Lewei Lu, Bin Li, Xiaogang Wang, and Jifeng Dai.
\newblock Deformable detr: Deformable transformers for end-to-end object
  detection.
\newblock {\em ICLR}, 2021.

\end{thebibliography}
}

\end{document}